# Argumentation in Waltz's "Emerging Structure of International Politics"

MAGDALENA WOLSKA [*,1], BERND FRÖHLICH [*], KATRIN GIRGENSOHN [†], SASSAN GHOLIAGHA [†], DORA KIESEL [*], JÜRGEN NEYER [†], PATRICK RIEHMANN [*,‡], MITJA SIENKNECHT [†], BENNO STEIN [*]

[*] *Bauhaus-Universität Weimar*
*Weimar, Germany*
{magdalena.wolska,bernd.froehlich,dora.kiesel,benno.stein}@uni-weimar.de

[†] *Europa-Universität Viadrina*
*Frankfurt (Oder), Germany*
{girgensohn,gholiagha,neyer,sienknecht}@europa-uni.de

[‡] *Jönköping University*
*Jönköping University*
patrick.riehmann@ju.se

ABSTRACT: We present an annotation scheme for argumentative and domain-specific aspects of scholarly articles on the theory of International Relations. At argumentation level we identify Claims and Support/Attack relations. At domain level we model discourse content in terms of Theory and Data-related statements. We annotate Waltz's 1993 text on structural realism and show that our scheme can be reliably applied by domain experts enables insights on two research questions on justifications of claims.

KEYWORDS: annotation, argumentation, political discourse

## 1. INTRODUCTION

Over the past years there has been a growing interest in the analysis of the language and discourse of politics. Numerous studies have focussed on the analysis of various aspects of political discourse, inluding modelling political debates (Villares and He, 2017; Haddadan et al, 2019; Padó et al, 2019; Goffredo et al, 2022; Mancini et al, 2022), creation of corpora such as the DCEP (Hajlaou et al, 2014) or JRC-Acquis (Steinberger et al, 2006) and tagged corpora of parliamentary debates (see, for instatnce, (Abercrombie and Batista-Navarro, 2018; Abercrombie and Batista-Navarro, 2020)), analysis of specific political speeches (Beelen et al, 2017; Labbé and Savoy, 2021; Card et al, 2022), or analysis of higher-level pragmatic aspects such as bias (Fischer-Hwang et al, 2022; Davis et al, 2022), manipulation, and politeness (Abuelwafa, 2021; Moghadam and Jafarpour, 2022; Kádár and Zhang, 2019; Trifiro et al, 2021).

While most prior research into the universe of political discourses is based in the genres of debate and speeches, studies of academic political discourse have been sparse. One of the goals of the project SKILL, from which this paper stems, is to fill this gap. SKILL – A social science lab for research-based learning – is dedicated to building and applying AI technologies to facilitate analysis of argumentation in scholarly articles in political science, especially in the context of teaching International Relations (IR). The ultimate goal of SKILL is to provide students with AI tools which would facilitate comprehension of original articles used as part of teaching syllabi and which would coach them in producing expert argumentation in the field.

In order to gain insight into the structure and properties of arguments in the domain of political science theory, we developed an annotation scheme which enables analysis of scholarly IR discourse in terms of interaction between argumentation and types of domain content contributing to arguments. The scheme comprises two orthogonal dimensions: discourse and content domain. At the discourse dimension, we model argumentation (using a basic model of premise-conclusion structures) and basic rhetorical structure (identifying elaborative discourse segments). At the domain dimension we focus on contributions relevant from the point of view of the domain of International Relations and distinguish between theoretical statements, definitions, and two types of empirical statements, while allowing for other types, not explicitly named at the time of scheme development. We apply the scheme to Waltz's 1993 text and address two research questions relevant from the point of view of teaching IR based on this source:

> **RQ1**: To what extent is evidence for claims explicitly provided in Waltz's text?
> **RQ2**: Is Waltz's argumentation mainly grounded in theory or in empirics?

---
1   Corresponding author.

Figure 1. Overview of the annotation scheme; "None" are technical categories indicating no annotation at the given level.

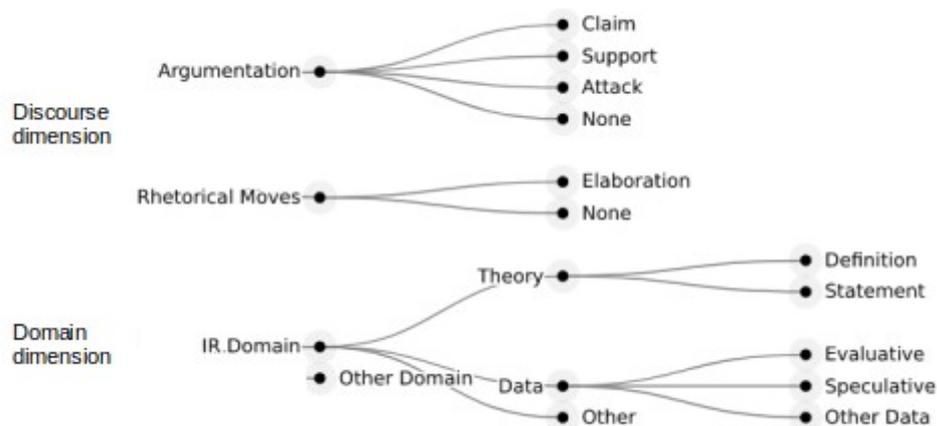

The paper is structured as follows: We start by presenting our model of theoretical discourse in IR in Section 2. In Section 3 we outline our data, the scheme development process, and the annotation procedure. In Section 4 we present an analysis of premise-conclusion structures addressing our research questions and summarize our contributions in Section 5.

2. MODELLING ARGUMENTATION IN SCHOLARLY DISCOURSE ON INTERNATIONAL RELATIONS

In order to investigate argumentation in scholarly discourse in political science, we developed an annotation scheme which models, on the one hand, basic discourse structure, and, on the other hand, the types of argumentative content relevant in our domain of discourse. Our annotion scheme comprises two dimensions, Discourse and Domain. The Discourse dimension describes argumentation and the rhetorical structure and can be applied to any argumentative text. The domain dimension describes the discourse contributions in terms of the type of content they present, that is, types of discourse moves specific to the domain of discourse; in our case, the theory of International Relations. The overview of the annotation scheme is shown in Figure 1. Annotation categories within the two dimensions are defined below.

DISCOURSE DIMENSION

The Discourse dimension models argumentation and rhetorical aspects of text. At the level of *Argumentation* we model discourse structures which build up an argument, that is, we identify those discourse moves that contribute to bringing argumentation forward as well as relations between those moves. Our argumentation-related categories are a simplified set of argumentative moves proposed by Toulmin (1958). The original Toulmin model of argumentation has been widely used in studies of argumentative discourse, however, it has been shown to present difficulties for annotation of real life argumentation (see, for instance, (Simosi, 2003)). Argumentation has been also shown to be difficult to annotate in general, yielding low interannotator agreement (see, for instance, (Torsi and Morante, 2018)). We therefore opt to model argumentation at the lowest level of complexity, namely, by only identifying basic premise-conclusion structures in terms of Claims and two relations that may hold between them, Support and Attack, defined as follows:[2]

> **Claim** is a statement that presents a basic building block of an argument. It is the assertion that a party puts forth and would like to convince the audience of, that is, prove. A claim can be also thought of as the conclusion that a party in discourse is attempting to draw.
>
> **Support** in an argument is a statement that provides evidence justifying a claim. This may be a statement that directly brings up facts, data, or other pieces of evidence showing why a claim holds. The purpose of a supporting statement is to increase the credibility of a claim, i.e. the readers' belief that the claim holds.
>
> **Attack** is a counter-argument to a previously proposed claim. The purpose of an attacking statement is to decrease the credibility of a claim, i.e. the readers' belief that the claim holds.

Note that unlike other argumentation annotation schemes (e.g. (Stab & Gurevych, 2014)) we do not distinguish between so-called main/major and minor claims at this point. Since our data comprises research articles, i.e., longer discourses of high linguistic complexity, we again opt for refraining from adding to the complexity of annotation. However, we approximate the distinction between major and minor claims by modeling local elaboration structures at the rhetorical level explicitly (see below). Discourse units which are not argumentative in the sense of the three categories defined above remain unlabelled at the argumentative level.

At the level of ***Rhetorical Moves*** we model the structural organization of text, i.e. the rhetorical roles of spans of text in a larger discourse which make the discourse coherent. Depending on a linguistic theory, rhetorical phenomena in discourse may encompass up to even 30 types of rhetorical coherence types (Taboada & Mann, 2006) including relations such as Background (facilitates understanding), Evaluation (evaluative comment), Purpose (intent behind a state or action), Means (method or instrument that facilitates realization of an action). Note that argumentation itself is also a rhetorical phenomenon which can be modelled at finer detail than the Claim-Support/Attack model presented above, using rhetorical relations such as Evidence, Explanation, and (volitional and non-volitional) Cause and Result. We model argumentation as a distinct level of annotation since it pays a central role in our model and we focus directly on basic argumentative premise-conclusion structures. At the rhetorical level we annotate a single relation, Elaboration, defined as follows:

> **Elaboration** expands on a point by contextualizing it or provides more information about a previous statement. It may describe it in a different way (e.g. restate, paraphrase, or reformulate it) or at a different level of abstraction (e.g. make it more specific/general)

Elaboration as defined above combines presentational aspects (cf. Mann and Thompson's (1988) Reformulation/Restatement and Summary and Hobbs' (1979) Repetition) as well as content aspects (cf. Danlos' (1999) Particularization and Generalisation) of a discourse unit. The main purpose of Elaboration in our scheme is to facilitate setting apart main claims from minor (elaborated) claims in argumentation. Statements which are not elaborative remain unannotated at the Rhetorical Level.

DOMAIN DIMENSION

The types of content contributed to discourse depends on the discourse genre and, naturally, on the domain of discourse. For instance, in the medical domain there might be discourse contributions related to a patient's diagnosis, in the music domain to the structure of a musical piece, and in the domain of chemistry to the interactions between chemical elements. In our case of political science domain, the domain dimension models the type of content specific to presenting a political science theory, in particular, theory of International Relations. For statemens within the ***IR Domain***, that is those about International Relations or global politics, we distinguish between content related to ***Theory*** and ***Data*** with two subtypes each as defined below:

> ***Theory*** statements present theoretical postulates. Empirical references or illustrations may be also made within theory statements, however, as soon as a theoretical assertion is presented as a generalization going beyond any specific empirical references or illustrations, it should be annotated within the Theory category. Two subcategories are explicitly defined:
>
>> **Definition** is a statement which explicitly specifies a meaning of a term used in the domain.
>>
>> **Theoretical Statement** is a non-definitional theoretical statement, i.e. one which is about IR-relevant theoretical concepts or topics.
>
> ***Data*** statements provide relevant empirical evidence, i.e. concrete reference to the real world, including classes of events (e.g. war), or social facts. Two subcategories are explicitly defined:
>
>> **Speculative Empirical Reference** makes a statement about a possible present or future scenario or an alternative past scenario; neither of those has actually happened.
>>
>> **Evaluative Statement** contains a reference to real world events, data, or (social) facts which are evaluated or interpreted by the author from any theoretical standpoint or presents it as a fact through a theory's perspective.
>
> Any other statements about real world annotated as **Other Data**.

---

[2] We use capitalized "Claim" to refer to a markable of type Claim and lower-case "claim" when we talk about argumentative claims in general.

In case a statement does present IR relevant content, but it cannot be classified as Theory or Data according to definitions above, we annotate it as **Other**, which makes the scheme open-ended at the domain dimension. If a statement explicitly refers to a domain other than political science, International Relations, or global politics, it is annotated as **Other Domain**. Figure 1 shows the category structure of the annotation scheme and Figure 2 illustrates the categories on an excerpt from (Waltz, 1993).

## 3. MATERIALS AND METHODS

### 3.1 LINGUISTIC DATA

As a basis for annotation scheme development and validation we selected foundational texts introducing four major theories of International Relations: neorealism (Waltz's 1993 "The Emerging Structure of International Politics"), liberalism (Putnam's 1988 "Diplomacy and Domestic Politics: The Logic of Two-Level Games"), constructivism (Finnemore's and Sikkink's 1998 "International Norm Dynamics and Political Change"), and feminism (Carpenter's 2005 "`Women, Children and Other Vulnerable Groups': Gender, Strategic Frames and the Protection of Civilians as a Transnational Issue"). Only the body of the articles – without footnotes and endnotes – was analyzed and annotated.

The articles were prepared for analysis by segmenting into sentences in a semi-automatic process. A sentence was defined, in a standard fashion, as a linguistic unit which expresses a complete thought and typically consists of a subject and predicate. Aside from the typical end-of-sentence punctuation (full-stop, question mark, and exclamation mark), sentence boundaries were also identified by semicolons, colons and (em) dashes, which are often used in scholarly articles to delimit parts of a sentence which could also be rendered as separate stand-alone sentences. Sentence segmentation was checked and corrected manually by one of the co-authors with linguistic background.

## 3.2 SCHEME DEVELOPMENT

The annotation scheme presented in Section 2 was developed in a combined theory-driven and data-driven fashion as a collaboration between senior scholars in International Relations and Linguistics, all co-authors of this paper.

The Discourse dimension was derived from preexisting approaches to rhetorical structure and argumentation analysis. Key modifications to an existing argument modeling scheme involved simplification (see Section 2). While the initial set of rhetorical functions included also, for instance, (rhetorical) Questions and Quotations, we ultimately did not include them in the final scheme since these categories can be reliably identified computationally even using just heuristics.

The Domain dimension was created in several iterations by analyzing excerpts from the four theory-foundational articles mentioned in Section 3.1 and applying tentative variants of the scheme to different fragments of the articles than the ones used for scheme development. Tentative schemes at the Domain level included variants with more fine-grained categories and with alternative definition wording. For example, in one of the early variants of the scheme, we differentiated between tree different types of Theory statements: Foundational statements, Assumptions, and Inferrences. Foundational statements were meant as building blocks shared between (some) IR theories, Assumptions as statements laying out a specific theory's premises and being taken for granted, and Inferences as derivable from either Foundational statements or Assumptions. However, the distinction between Assumptions and Inferences proved difficult to pinpoint rigorously, which led to a large number of disagreements between annotators. Ultimately, we arrived at a scheme which is a compromise between reliability, cost, and descriptive power: our annotators reach satisfactory agreement on a model that targets research questions of our interest.

## 3.3 ANNOTATION PROCEDURE

Annotation was conducted using dedicated annotation software developed specifically for the purpose of the project. Discourse annotation was done within a scope of a paragraph, that is, no cross-paragraph argumentative and rhetorical relations are identified.

Within each paragraph, annotators proceed as follows: First, they read the entire paragraph for comprehension, register its overall message and the flow of argumentation. Second, they proceed sentence by sentence and identify all claims within the paragraph by first identifying the sentence's predicate-argument structure and then answering the question whether its content is relevant from the point of view of political science, especially International Relations, or global politics, i.e. whether the sentence is about the IR Domain. If so, then if the sentence contributes argumentative content, i.e. it brings the thread of argumentation forward, it is marked as Claim and Rhetorical and Domain categories are assigned according to definitions. Once this is done, for each Claim its Supporting and Attacking claims (if any) within the given paragraph are identified.

## 3.4 ANNOTATORS AND CODING DISAGREEMENTS

For the purpose of the work presented here annotations were performed by two domain experts, co-authors of this article. Both domain experts are senior researchers in political science whose focus of research and academic teaching is on International Relations. One domain expert has had somewhat less experience coding with the final version of our annotation scheme, but trained in using it prior to starting annotating the Waltz's text. Annotation was first performed by each of the domain experts independently. Once the entire text was annotated, the annotators met, discussed discrepancies, and agreed on final annotations.

Table 2. Basic text descriptives on Waltz's "The Emerging Structure of International Politics"; number of words shown excludes punctuation

| Feature | Value |
| --- | --- |
| Number of sentences | 641 |
| Number of words | 13064 |
| Number of distinct words | 2768 |

The majority of disagreements at the Disocurse dimension were at the Argumentation level and were due to unidentified argumentative relations by one of the annotators. There were 140 such cases where one of the experts has interpreted a relation between Claims, either Support or Attack, whereas the other did not see one. Most of those disagreements, however, can be attributed to unequal experience of the annotators in using the scheme since they were eventually settled in line with the interpretation of the annotator who had been working with the scheme for a longer time. There were also 7 cases of disagreements where one of the experts interpreted a Support relation and the other an Attack. We will investigate those inconsistent relation interpretations further as we work with other texts.

There were 159 disagreements at the Domain dimension (25% disagreement rate) of which 94 were between main categories within the Domain dimension, i.e. Theory vs. Data. The error in assigning Data over Theory was systematic and we attribute it to oversight since the instructions for coding the Theory category were modified several times in the course of the annotation scheme development. An example of such annotation artefact is the segment "In the nuclear era, countries with smaller economic bases can more easily achieve great-power status." which one of the experts annotated as Data→Evaluative and the other as Theory→Statement. In the context of the preceding sentences, the segment can be considered a borderline case: As such, no specific real world events are mentioned here and the statement is a generic one, however, from prior context it is clear which smaller economies are meant. In the course of guidelines development, we adjusted emphasis on theoretical content, which was overlooked on cases like this one.

For each of the disagreement cases domain experts were able to reach consensus without resorting to adjudication by a third expert. The resulting consensus annotations, to which we refer to as "gold standard", was used for the analysis.

## 4. ARGUMENTATION AND DOMAIN DISCOURSE IN WALTZ'S "THE EMERGING STRUCTURE OF INTERNATIONAL POLITICS"

Basic descriptive information about Waltz's text used this in study is shown in Table 2. Since annotation was performed at the sentence level, the number of sentences here, 641, corresponds also to the number of annotated segments. We start the analysis with general observations about Waltz's argumentation and then characterize it in terms of the types of domain content Waltz uses in his arguments. For this overview we use the Gold Standard annotations, that is, the annotation agreed on by both domain experts upon discussion.

### 4.1 PREMISE-CONCLUSION STRUCTURES

Out of the 641 annotated segments in Waltz's article, 603 are part of argumentation (i.e. they have been annotated as Claim, Support, or Attack). Examples of Claim-Support and Claim-Attack links are shown in Figure 2 in an excerpt of Waltz's text as displayed in the annotation software. Waltz's text is thus for the most part argumentative with only about 5% of the sentences not having directly argumentative function. The segments which do not constitute part of argumentation are mainly rhetorical questions (such as "Why should the future be different from the past?", "What are the possibilities?"), references to the text outline or structure (such as, "The next section asks what differences this will make in the behavior and interaction of states.", "The preceding paragraph reflects international-political reality through all of the centuries we can contemplate."), and other non-argumentative segments (e.g. "Two points about nuclear weapons remain.", "The case of Western Europe remains.").

Figure 2. An except of Waltz's text in the annotation software with example annotations. Discourse dimension categories are shown in the left margin of the text; Domain dimension is shown on the right. Arrows on the left denote argumentative links: Claim-Support (solid lines) and Claim-Attack (wavy lines)

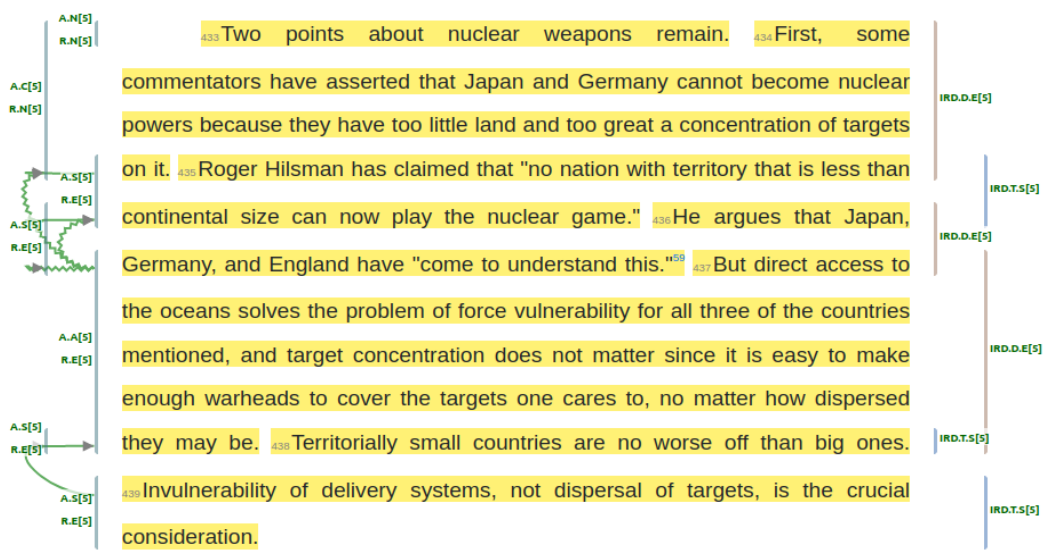

The majority of Waltz's Claims, 440, are part of elaborated structures and only 163 – 27% of all the Claims and 25% of all the segments – are what can be considered "main" or "major" claims, i.e. they possibly initiate an elaborated segment and can be thought of as Waltz's core train of reasoning. Indeed, among those we find statements such as "The conflation of peace and stability is all too common.", "Structural change begins in a system's unit, and then unit-level and structural causes interact.", "The political and economic reconstruction attempted by the Soviet Union followed in part from external causes.", or "But great powers do not gain and retain their rank by excelling in one way or another."

In order to answer our first research question as to overt justification for claims (see RQ1 in Section 1), we looked into the proportion of supported and unsupported claims. Only 156 out of the 603 Claims (26%) in Waltz's text are provided with supporting evidence within text, that is, they form Claim-Support chains and can be considered arguments in the sense of Premise-Conclusion structures. The majority of those, 116, are provided with a single evidence statement, whereas a small proportion, 6, have three, four, and even 6 supporting statements. The remaining 447 claims are presented without evidence in the text, i.e. there is no other claim which stands in a Support relation to those claims. It can be assumed then that the author considers justification for these claims to be part of the so-called *common ground* or shared understanding/common knowledge, i.e. "general knowledge shared by the speaker, hearer, and audience" (Walton 1996; see also Clark and Schaefer, 1989; van Eemeren & Grootendorst, 2004). In itself, the fact that many claims are provided without support is not surprising. However, it has implications on teaching the neorealist theory of International Relations and on its understanding by the students based on this article, in that the background knowledge assumed by Waltz needs to be addressed by the instructor.

4.2 TYPES OF DOMAIN CONTRIBUTIONS IN PREMISE-CONCLUSION STRUCTURES

In order to address the second research question of whether Waltz's argumentation is mainly grounded in theory or in empirics (see RQ2 in Section 1), we analyze the distribution of Theory and Data categories within Claims and Supports in Waltz's text. Out of the 603 Claims a was majority, 406, is in the Data category. There are 194 Theory claims, one claim from a non-IR domain (unannotated at the Domain level) and three claims with domain categories which we did not anticipate (annotators marked those as "Domain: Other"). In general thus, Waltz's argumentation in mainly empirically driven.

Most "major" claims – i.e. those which are not part of elaboration structures – are also Data-oriented. There are 90 Data-oriented major claims in total: 68 Evaluative major claims about actual events in the world and 21 Speculative claims about future or past scenarios. 61 major Claims are Theory Statements. Analogous distribution can be found within supporting evidence: out of the 235 Support claims, 170 (72%) are grounded in Data and only 64 (27%) in Theory.

The overall picture that emerges is that Waltz's theory-oriented paper is for the most part driven by and grounded in empirics. From an educational perspective this means that the key prerequisites for comprehending Waltz's argumentation are strong background in history and an ability to recognize the impact of world events on international relations, of which instructors who use Waltz's text as part of undergraduate syllabi should be made aware.

5. CONCLUSION

We presented an annotation scheme developed for the purpose of studying argumentation and domain discourse as well as their interaction in scholarly articles in political science. We have shown that our scheme can be reliably applied to annotate Waltz's "The Emerging Structure of International Politics", one of the fundational texts on the neorealist theory of International Relations used in academic Political Science programmes. Based on the annotated text we have also shown that Waltz's argumentation is grounded mainly in empirics and that most of his claims are not explicitly justified within the text. In the context of using Waltz's article in undergraduate curricula this poses strong assumptions and requirements on students' understanding of the historical background, both in terms of knowledge of historic events themselves as well as their political impact. Our further work will concentrate on building a larger annotated corpus of reseach articles on International Relations and further, larger scale corpus-driven analysis of the relation between argumentation and domain discourse based on this data.

ACKNOWLEDGEMENTS: Work described in this paper is part of the SKILL project, "Sozialwissenschaftliches KI-Labor für Forschendes Lernen" (A social science lab for research-based learning), funded by the German Ministry for Education and Research, the Brandenburg Ministry for Science and Culture and the Thuringian Ministry for Science, Research and Art.

REFERENCES

Abercrombie, G., & Batista-Navarro, R. (2018). A sentiment-labelled corpus of hansard parliamentary debate speeches. In Proceedings of ParlaCLARIN. Common Language Resources and Technology Infrastructure (CLARIN).
Abercrombie, G., & Batista-Navarro, R. T. (2020). Parlvote: A corpus for sentiment analysis of political debates. In Proceedings of the 12th Language Resources and Evaluation Conference.
Abuelwafa, M. A. (2021). Legitimation and manipulation in political speeches: a corpus-based study. Procedia Computer Science, 189.
Beelen, K., Thijm, T. A., Cochrane, C., Halvemaan, K., Hirst, G., Kimmins, M., and others (2017). Digitization of the Canadian parliamentary debates. Canadian Journal of Political Science/Revue canadienne de science politique, 50(3).

Card, D., Chang, S., Becker, C., Mendelsohn, J., Voigt, R., Boustan, L., and others (2022). Computational analysis of 140 years of us political speeches reveals more positive but increasingly polarized framing of immigration. Proceedings of the National Academy of Sciences, 119(31).

Clark, H. H., & Schaefer, E. F. (1989). Contributing to discourse. Cognitive Science, 13(2).

Danlos, L. (2001). Event coreference between two sentences. Computing Meaning: Volume 2.

Davis, S. R., Worsnop, C. J., & Hand, E. M. (2022). Gender bias recognition in political news articles. Machine Learning with Applications, 8.

Fischer-Hwang, I., Grosz, D., Hu, X. E., Karthik, A., & Yang, V. (2020). Disarming loaded words: Addressing gender bias in political reporting. In Computation + Journalism Conference.

Goffredo, P., Haddadan, S., Vorakitphan, V., Cabrio, E., & Villata, S. (2022). Fallacious argument classification in political debates. In Proceedings of the 31st International Joint Conference on Artificial Intelligence.

Haddadan, S., Cabrio, E., & Villata, S. (2019). Yes, we can! Mining arguments in 50 years of us presidential campaign debates. In Proceedings of the 57th Annual Meeting of the Association for Computational Linguistics.

Hajlaoui, N., Kolovratnik, D., Väyrynen, J., Steinberger, R., Varga, D., et al. (2014). DCEP – Digital Corpus of the European Parliament. In Proceedings of the 9th Language Resources and Evaluation Conference.

Hobbs, J. R. (1979). Coherence and coreference. Cognitive science, 3(1).

Kádár, D. Z., & Zhang, S. (2019). (im) politeness and alignment: A case study of public political monologues. Acta Linguistica Academica, 66(2).

Labbé, D., & Savoy, J. (2021). Stylistic analysis of the french presidential speeches: Is macron really different? Digital Scholarship in the Humanities, 36(1).

Mancini, E., Ruggeri, F., Galassi, A., & Torroni, P. (2022). Multimodal argument mining: A case study in political debates. In Proceedings of the 9th Workshop on Argument Mining.

Mann, W. C., & Thompson, S. A. (1988). Rhetorical structure theory: Toward a functional theory of text organization. Text-interdisciplinary Journal for the Study of Discourse, 8(3).

Moghadam, M., & Jafarpour, N. (2022). Pragmatic annotation of manipulation in political discourse: The case of Trump-Clinton presidential debate. Linguistic Forum – A Journal of Linguistics, 4(4).

Padó, S., Blessing, A., Blokker, N., Dayanık, E., Haunss, S., & Kuhn, J. (2019). Who sides with whom? towards computational construction of discourse networks for political debates. In Proceedings of the 57th Annual Meeting of the Association for Computational Linguistics.

Simosi, M. (2003). Using toulmin's framework for the analysis of everyday argumentation: Some methodological considerations. Argumentation, 17(2).

Stab, C., & Gurevych, I. (2014). Annotating argument components and relations in persuasive essays. In Proceedings of the 25th International Conference on Computational Linguistics: Technical papers.

Steinberger, R., Pouliquen, B., Widiger, A., Ignat, C., Erjavec, T., Tufis, D., & Varga, D. (2006). The JRC-Acquis: A multilingual aligned parallel corpus with 20+ languages. In Proceedings of the 5th Language Resources and Evaluation Conference.

Taboada, M., & Mann, W. C. (2006). Rhetorical structure theory: Looking back and moving ahead. Discourse Studies, 8(3).

Torsi, B., & Morante, R. (2018). Annotating claims in the vaccination debate. In Proceedings of the 5th Workshop on Argument Mining.

Toulmin, S. E. (2003). The uses of argument. Cambridge University Press.

Trifiro, B. M., Paik, S., Fang, Z., & Zhang, L. (2021). Politics and politeness: Analysis of incivility on twitter during the 2020 democratic presidential primary. Social Media + Society, 7(3).

Van Eemeren, F. H., Grootendorst, R., & Grootendorst, R. (2004). A systematic theory of argumentation: The pragma-dialectical approach. Cambridge University Press.

Vilares, D., & He, Y. (2017). Detecting perspectives in political debates. In Proceedings of the 2017 Conference on Empirical Methods in Natural Language Processing.

Walton, D. N. (1996). Argument structure: A pragmatic theory. University of Toronto Press.